\def\year{2021}\relax
\documentclass[letterpaper]{article} 
\usepackage{aaai21}  
\usepackage{times}  
\usepackage{helvet} 
\usepackage{courier}  
\usepackage[hyphens]{url}  
\usepackage{graphicx} 
\usepackage{natbib}  
\usepackage{caption} 
\usepackage{amsmath,amsfonts,bm}

\def\eqref#1{equation~\ref{#1}}

\def\1{\bm{1}}

\DeclareMathAlphabet{\mathsfit}{\encodingdefault}{\sfdefault}{m}{sl}
\SetMathAlphabet{\mathsfit}{bold}{\encodingdefault}{\sfdefault}{bx}{n}

\newcommand{\R}{\mathbb{R}}

\usepackage{hyperref}
\usepackage{url}
\usepackage{graphicx}
\usepackage{wrapfig}
\usepackage{xcolor}
\newcommand{\com}[1]{{\color{red}#1}}
\usepackage{comment}
\usepackage{enumitem}
\usepackage{textcomp}
\usepackage{multirow}
\usepackage{colortbl}

\usepackage{amsfonts}
\usepackage{booktabs}
\usepackage{siunitx}

\usepackage{amsmath}
\usepackage{amssymb}

\usepackage[capitalize,noabbrev]{cleveref}

\renewcommand{\ref}{\cref}

\newcommand{\N}{\mathbb{N}}

\newcommand{\mr}[1]{\mathrm{#1}}
\newcommand{\D}{\mathcal{D}}
\newcommand{\T}{\mathcal{T}}

\title{Selective Forgetting  of Deep Networks at a Finer Level than Samples}

\author{
    Tomohiro Hayase,
    Suguru Yasutomi,
    Takashi Katoh\\
} \affiliations{
    Fujitsu Laboratories Ltd. \\
    Kawasaki, Kanagawa, Japan \\
    hayase.tomohiro@fujitsu.com,
    yasutomi.suguru@fujitsu.com,
    kato.takashi\_01@fujitsu.com
}

\begin{document}

\maketitle

\begin{abstract}
Selective forgetting or removing information from deep neural networks (DNNs) is essential for continual learning and is challenging in controlling the DNNs.
Such forgetting is crucial also in a practical sense since the deployed DNNs may be trained on the data with outliers, poisoned by attackers, or with leaked/sensitive information.
In this paper, we formulate selective forgetting for classification tasks at a finer level than the samples' level. 
We specify the finer level based on four datasets distinguished by two conditions: whether they contain information to be forgotten and whether they are available for the forgetting procedure.
Additionally, we reveal the need for such formulation with the datasets by showing concrete and practical situations.
Moreover, we introduce the forgetting procedure as an optimization problem on three criteria; the forgetting, the correction, and the remembering term.
Experimental results show that the proposed methods can make the model forget to use specific information for classification.
Notably, in specific cases, our methods improved the model's accuracy on the datasets, which contains information to be forgotten but is unavailable in the forgetting procedure.
Such data are unexpectedly found and misclassified in actual situations.
\end{abstract}

\begin{figure*}[t]
    \centering
    \includegraphics[width=\linewidth]{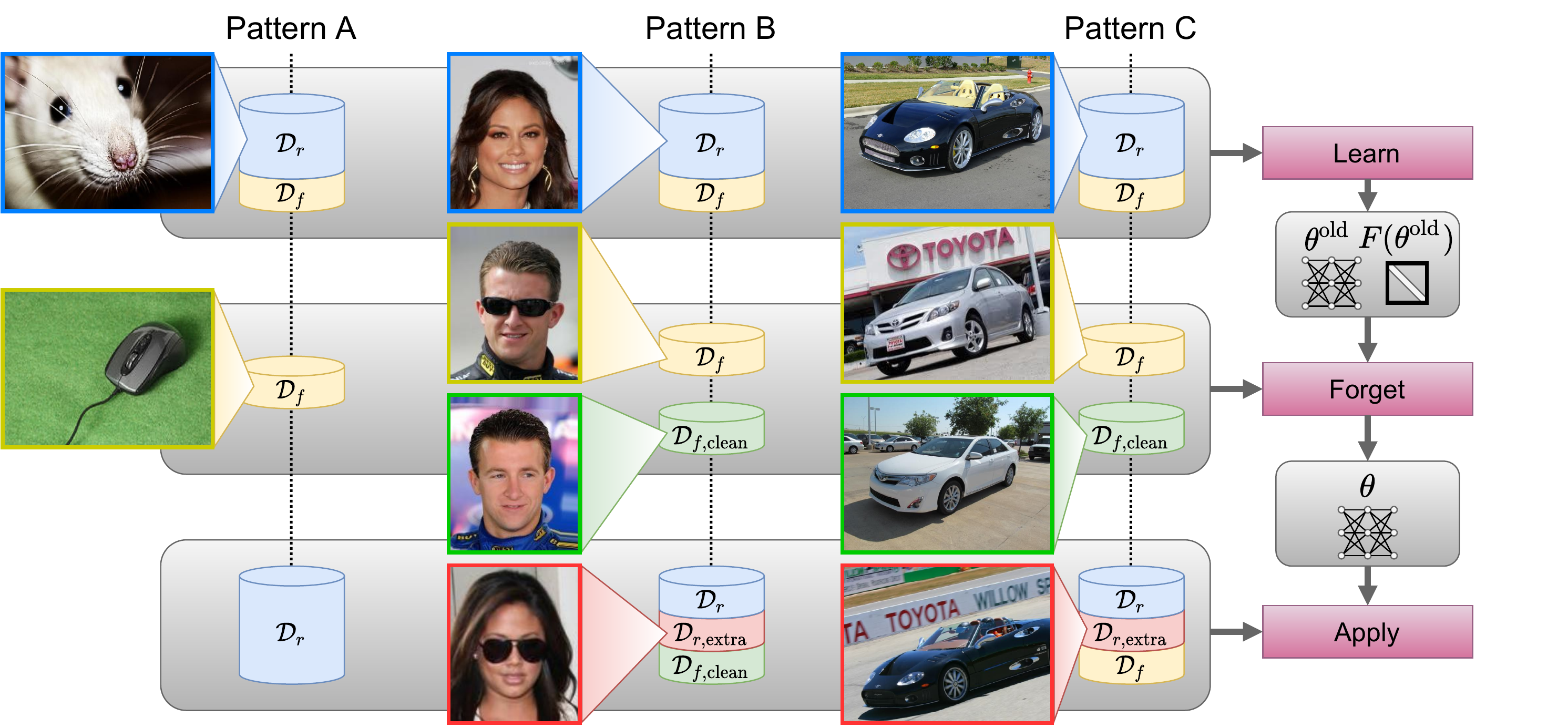}
    \caption{%
        Overview of our approach.
        We formulate selective forgetting by four datasets; $\D_r$, $\D_f$, $\D_{f, \text{clean}}$, and $\D_{r, \text{extra}}$.
        Depending on which dataset the DNN should perform well, three patterns of forgetting are introduced.
        In Pattern A, the DNN forgets samples such as outliers.
        As illustrated in the figure's left side, images of mice that are input devices are outliers when the task is classifying animals, for example.
        In Pattern B, the DNN forgets a part of each training sample.
        The model is supposed to forget to use the sunglasses as a feature for classification in the figure.
        Pattern C is similar to Pattern B; in the figure, the model is supposed to forget to use the logo or the emblem of a carmaker for classifying types of cars.
        Pattern B and C are different in their evaluation.
        Example images are from WebVision~\citep{li2017webvision}, CelebA~\citep{liu2015faceattributes}, and Cars Dataset~\citep{KrauseStarkDengFei-Fei_3DRR2013}.
        }
    \label{fig:overview}
\end{figure*}

\section{Introduction}
In practical applications of machine learning, models must deal with continually arriving input data.
Lifelong machine learning~\citep{Chen2018,PARISI201954} is a framework addressing this problem.
It consists of various techniques, such as continual learning, transfer learning, meta learning, and multi-task learning.
Continuous learning is the most straightforward idea of lifelong machine learning.
It aims to accumulate knowledge to a model from many tasks and data which come intermittently. 
Eventually, we expect the model to solve different types of tasks on a wide range of data.
However, indeed, there are many difficulties in achieving such kind of learning.

In terms of deep neural networks (DNNs), the most typical problem on continual learning is catastrophic forgetting~\citep{kirkpatrick2017overcoming,Li2018}.
If we train an already trained DNN on a new task, the parameters of the DNN will be overwritten, and will completely forget the previous task.
This behavior is called catastrophic forgetting.
Previously proposed techniques can alleviate this problem and have shown the possibility to add information to DNNs~\citep{kirkpatrick2017overcoming,kemker2018measuring} continually.

Selective forgetting, which is a subtraction of information, for DNNs is also crucial for continuous learning.
In practical situations, training data often contain useless or undesired data, and we may want to remove such information from the model afterward.
Especially in industrial scenes, various problems can appear in long-term operation even if developers thought everything was fine at the first stage of deployment.

Typically, a trained DNN is required to forget specific samples in the training dataset.
One reason for the request is the poor performance of inference caused by outliers.
If the dataset has noisy outliers, the model's generalization performance gets low.
Privacy, which is related to GDPR (General Data Protection Regulation) in Europe and the right to be forgotten, is also a popular reason.
For example, when you construct an image dataset using images on the Internet and train a DNN with it, some right holders of the images may demand removing their information from both the dataset and the trained model.
From a privacy-protection perspective, selective forgetting is quite difficult because the training data could be estimated from the model~\citep{Fredrikson2015}. 
Continuously learning DNNs such as a chatbot need selective forgetting, of course.
Chatbots often learn from users' posts, and its corpus is frequently polluted.
Rolling back is a possible solution, but it removes both recent useful and useless corpora.
It can be a better solution to forget only the useless corpus selectively and reserve the useful corpus.

Targets for selective forgetting can be a finer level than samples in many situations.
Poisoning~\citep{Munoz2017}, an attack that pollutes training data and makes models malfunction, can make one of the situations.
\citet{chen2017targeted} have illustrated that attackers can set up backdoors by injecting specific image patches to some training samples.
More concretely, by adding face images with specific glasses to the training data, the attackers can make a face recognizing DNN classify face images with the glasses as a class.
In such cases, it is desirable to make the DNN forget to use feature corresponding to the glasses rather than forget whole poisoned samples.
Leakage~\citep{Kaufman2012} or shortcut learning~\citep{Geirhos2020a} can also be situations that need forgetting.
If some of the training data contain something like data ID, the DNN exploits it, and the generalization performance would be ruined.
Hence, as in the case of poisoning, it is important to forget the leakage's effect.
In a context of fairness~\citep{binns18a}, some of the explanatory variables can be sensitive (e.g.\,sexuality, address, and racial information), and the model would be required to forget them.

%
%

In this paper, we formulate selective forgetting using four datasets $\D_r, \D_f, \D_{f, \text{clean}}$, and $\D_{r,\text{extra}}$ that we get the idea from considering practical situations (see Section \nameref{sec:infromation-to-be-forgotten}).
The four datasets are distinguished by two conditions: whether they contain information to be forgotten and whether they are available for the forgetting procedure.
Additionally, we derive a novel forgetting procedure and show that it successfully make the DNN forget selectively.

We formulate three patterns of selective forgetting in classification tasks.
In the first pattern, we make the DNN forget samples that contain information to be forgotten.
In this pattern, we split a dataset into two datasets: a dataset $\D_f$ that consists of data with the information to be forgotten and a dataset $\D_r$ consists of the rest. 
In the second and third patterns, we adjust the targets to be forgotten in a finer level than the samples.
In the patterns, features in input data can be seen as backdoor or leakage
To evaluate the forgetting of such information, we use additional two datasets; $\D_{f, \text{clean}}$ and $\D_{r, \text{extra}}$.
The dataset $\D_{f, \text{clean}}$ is drawn from the similar distribution as $\D_f$, but each sample in $\D_{f, \text{clean}}$ is processed not to contain the information to be forgotten.
Conversely, the dataset $\D_{r, \text{extra}}$ is drawn from the similar distribution as $\D_r$, but each sample in $\D_{r, \text{extra}}$ is modified to have the information.
The second and the third pattern differ in which dataset the DNN should perform well (see \cref{tab:datasets-and-evaluations}).
We describe three patterns and their importance with concrete and practical situations that need forgetting (see Section \nameref{sec:situations}). 

We propose a forgetting procedure as training with a combination of a forgetting term, a correction term, and a remembering term.
The forgetting term is based on a random distillation.
The correction term and a remembering term are based on elastic weight consolidation (EWC)~\citep{kirkpatrick2017overcoming}; the first term is a classification loss on the additional data, and the second term is a regularization restricting the parameters' movement by employing Fisher information.
Regarding that the original training data is large and hardly accessible, we only use the dataset to be forgotten and its variant in the forgetting procedure.
Experimental results show that the proposed method can make the DNN forget the target information in certain situations.
Notably, we have found that our methods improve the performance on the data not shown in both the pretraining and the forgetting procedure.



\section{Related Work}
\paragraph{Catastrophic Forgetting}
To alleviate the catastrophic forgetting, \citet{kirkpatrick2017overcoming} proposed EWC.
EWC estimates which parameter is important for previously learned tasks by calculating diagonal Fisher information matrix (FIM) on the previous task.
Many other techniques have been proposed to prevent catastrophic forgetting~\citep{kemker2018measuring}.

\paragraph{Formulations of Selective Forgetting}
An important point for selective forgetting is how to define DNNs' forgetting state.
\citet{guo2019certified,Golatkar_2020_CVPR} defined the states based on differential privacy~\citep{dwork2008}.
Differential privacy is an idea that two models trained by the same algorithm should have (almost) the same parameters when one of the models is trained on some dataset and the other is trained on the dataset without a sample in the dataset.
In short, differential privacy guarantees that the removal of a sample in a dataset does not (or hardly) affect the resulting model.
Data deletion~\citep{ginart2019} has a similar concept as differential privacy; forgetting (or deletion) must result in the model that trained on the dataset without the data to be forgotten.
\citet{bourtoule2020machine} also employed a similar definition of forgetting and named it machine unlearning.
Certified data removal~\citep{guo2019certified} relaxes differential privacy by comparing the two models; the model trained without the sample to be forgotten and the model trained with it and made forget it.
\citet{Golatkar_2020_CVPR} aimed for more practical definition especially on DNNs;
target to be forgotten is relaxed to a certain subset of the dataset instead of a sample in a dataset.
Moreover, it allows the model parameters to perturb in order to remove the information of the subset to be forgotten.
These formulations are mainly on forgetting one or more samples.
We formulate selective forgetting in finer information than samples.
In our formulation, a DNN's forgetting state means that the behavior of the model does not change depending on whether a dataset contains information to be forgotten.

\paragraph{Features Finer than Samples}
As a finer level feature than samples, backdoor is well-known~\citep{li2020backdoor}.
Backdoor is hidden features that the attacker injects into the training data.
It leads the model to predict as the attackers want.
Defense methods against the backdoor attacks have been proposed~\citep{li2020backdoor}, but most of them must be applied before the training, in contrast to the forgetting which is an operation after the training.

\paragraph{Methods for Selective Forgetting}
For certified data removal~\citep{guo2019certified}, a forgetting method for linear classifiers is proposed.
The method is basically based on an additive noise to the loss function in the training time and Newton's method on the dataset without data to be forgotten.
It is applicable when the last layer of the DNN is a linear layer.
\citet{bourtoule2020machine} proposed SISA training that trains several models with disjoint subsets of the original training dataset.
The models trained with SISA training can efficiently forget certain samples under the condition that the whole the dataset is stored and available in the unlearning algorithm.
In contrast, our method targets the case that the access to the dataset is restricted.
Scrubbing~\citep{Golatkar_2020_CVPR} is a perturbation of the parameters; it randomly moves the parameters in a direction that scrubs the information of the data to be forgotten and does not affect the rest.
The direction is derived from FIM or using neural tangent kernel~\citep{golatkar2020forgetting}, for example. 
Our method also modify parameters using randomness in a more naive way than the scrubbing.
Additionally, \citet{ginart2019} treated a forgetting methods for $k$-means not for DNNs. 
They proposed a quantized variant of $k$-means as a clustering method that is robust to removing data.

\section{Formulation}
\subsection{Setting}
Let $f_\theta: X\to Y$ be a DNN, where $\theta$, $X$, and $Y$ are the parameters of the DNN, the input space, and the label space, respectively.
We assume that the DNN is trained on a classification task using a dataset $\D\subset X\times Y$  and a loss function $L(f_\theta, \D)$.
We call $\D$ the \emph{pretraining dataset} hereinafter to distinguish between the dataset for the classification task and the dataset for the training procedure of the forgetting.
As a result of the pretraining, we have a parameter $\theta_0$ that makes $L(f_{\theta_0}, \D)$ sufficiently small.


Let $\D_f \subset \D$ be a set of data that contain information to be forgotten.
Write $\D_r = \D\setminus \D_f$, which is a dataset to be remembered.
A trivial solution for selective forgetting is retraining using $\D_r$.
However, this strategy is not practical because $\D_r$ is often huge and the training takes a long time.
Besides, we sometimes do not have access to $\D_r$.
Thus, we assume that $|\D_r| \gg |\D_f|$ and $\D_r$ is basically not accessible in the forgetting procedure.

\subsection{Information to be Forgotten}\label{sec:infromation-to-be-forgotten}
A pattern of selective forgetting is to make the DNN forget the information that $\D_f$ contains but $\D_r$ does not. 
In other words, the DNN is required to forget samples in $\D_f$ and to remember those in $\D_r$.
We evaluate the forgetting by the accuracy on the datasets:
we say that the DNN forgets $\D_f$ if it keeps high accuracy for $\D_r$ and achieves low accuracy for $\D_f$.
\citet{Golatkar_2020_CVPR,golatkar2020forgetting} utilize the DNN trained only on $\D_r$ as the forgotten state.
For classification problems, the DNN in such a state should pass our forgetting criterion. 
By evaluating the forgetting using accuracy on the datasets, we can easily apply the method to a wide range of models.


Further, we introduce patterns of selective forgetting of subtle information than samples. 
Here, the subtle information to be forgotten is determined by $\D_f$ and an additionally given dataset $\D_{f, \text{clean}}$.
The additional dataset $\D_{f, \text{clean}}$ has data similar to these in $\D_f$ but do not contain the information to be forgotten.
For convenience, we introduce a map $\T \colon X \times Y \to X \times Y$ to describe and add the information to be forgotten. 
The map $\T$ is assumed to satisfy the following two conditions. 
Firstly, since $\D_{f, \text{clean}}$ describes the forgotten version of $\D_{f}$,  
we assume that $\D_f \subset \T(X \times Y)$ and $\D_{f,\text{clean}} \subset \T^{-1}(\D_{f})$.
Secondly, the distance between $\T(\D_{f, \text{clean}})$ and $\D_f$ is assumed to be sufficiently small.
In this situation, we evaluate the forgetting by the accuracy on four datasets:
$\D_f$,
$\D_{f, \text{clean}}$,
$\D_r$,
and an extra dataset $\D_{r, \text{extra}} := \T(\D_r)$. %
$\D_{r, \text{extra}}$ has similar data to $\D_r$, but samples in $\D_{r, \text{extra}}$ have the information to be forgotten.
$\D_{r, \text{extra}}$ may not exist in some cases, but it is necessary for evaluating the performance of the forgetting.

For which dataset the DNN should achieve high (or low) accuracy depends on the information to be forgotten as shown in \cref{tab:datasets-and-evaluations}.
Concrete applications for each pattern in \cref{tab:datasets-and-evaluations} are described in a later section. 
\begin{table*}[t]
    \centering
    \begin{tabular}{lccccl}
        \toprule
        &  \multicolumn{2}{c}{Testing data} & \multicolumn{2}{c}{Additional data} & \\
        \cmidrule(lr){2-3}\cmidrule(lr){4-5}
        Forgetting pattern &$\D_r$ & $\D_f$ & $\D_{f, \text{clean}}$ & $\D_{r, \text{extra}}$ & Examples to be forgotten  \\\midrule
        Pretrained state & $\uparrow$ & $\uparrow$   & N/A        & N/A     &            \\\midrule
        Pattern A        & $\uparrow$ & $\downarrow$ & N/A        & N/A   & Samples     \\
        Pattern B        & $\uparrow$ &              & $\uparrow$ & $\uparrow$ & Backdoor\\
        Pattern C        & $\uparrow$ & $\uparrow$   &            & $\uparrow$ & Leakage \\
        \bottomrule
    \end{tabular}
    \caption{Relationship between accuracy for the datasets and targets for forgetting. $\uparrow$ and $\downarrow$ respectively denotes high/low accuracy on corresponding pattern and dataset. Blank cell means we do not care the corresponding accuracy.}
    \label{tab:datasets-and-evaluations}
\end{table*}

Above, we described that $\D_{r, \text{extra}}$ is obtained afterward.
However, it is often found at first; misclassification of samples that are similar to these in $\D_r$ reveals the need of the forgetting and the information to be forgotten, for example.
In such case, we choose $\T$ so that $\D_{r, \text{extra}} \subset \T(\D_r)$ is satisfied.
Then, we obtain the dataset for the forgetting procedure by $\D_{f, \text{clean}} \subset \T^{-1}(\D_f)$.

\subsection{Loss Function for Forgetting}
We formulate a loss function for selective forgetting as a combination of forgetting term $L_f(f_\theta, \D_f)$, the correction term $L_c(f_\theta, \D_{f,\text{clean}})$, and the remembering term $R(f_\theta, f_{\theta^\text{old}})$.
Here,  the remembering loss approximates the KL divergence against the old model $f_{\theta^\text{old}}$, where $\theta^{\text{old}}$ is the parameter of the network before applying selective forgetting.
The loss function for the selective forgetting is a linear combination of the three terms with non-negative weights.
The weights are hyperparameters and decided by cross validation.
Recall that we do not use $\D_r$ or $\D_{r,\text{extra}}$ in the minimizing the selective forgetting loss.
However, we require validation sets $\D_r^\text{val}$ and $\D_{r,\text{clean}}^\text{val}$ in deciding the hyperparameters.
We describe the specific form of the losses in a later section.



\section{Situations}\label{sec:situations}
We list several situations that need selective forgetting.
They correspond to each pattern in \cref{tab:datasets-and-evaluations} and \cref{fig:overview}.
We also clarify concrete content of the datasets (e.g. $\D_r$, $\D_f$, and $\D_{r, \text{extra}}$) in the examples.

\subsection{Patten A: Forget Samples}
Generally, DNNs are good at dealing with large datasets which are costly.
To save the cost, we can use automatically collected datasets such as WebVision database~\citep{li2017webvision}.
Selective forgetting plays an important role in the DNNs' learning noisy and huge datasets like WebVision.
Suppose you have a DNN trained on WebVision and you found some outliers that affect the performance of the DNN.
The most naive way to remove the effect of the outliers is retraining without them.
However, the dataset is huge and it may take a couple of weeks to complete the training.
In this case, it is useful to forget the outliers in a short time without accessing whole the dataset.

In the context of continual learning, the need for selective forgetting is clearer.
Consider a chatbot that learns continually; the bot learns from users' reactions.
Even if the bot is once successfully trained on useful information, malicious users may teach irrelevant expressions.
Since the bot is in continual learning, it will soon make such expressions like Microsoft Tay which ended up repeating racist remarks because of the poisoned corpus caused by malicious users~\citep{neff2016automation,WOLF20171}.
Rolling back the bot to the state before the attack is a trivial solution in such a case.
However, the bot may learn proper expressions by normal users during the attack.
If we can make the bot forget bad corpus and preserve the others, the bot will be under control and can continue to work after the attack.

Such cases require the DNN of forgetting specific samples.
They correspond to Pattern A in \cref{tab:datasets-and-evaluations}:
the DNN must achieve low accuracy on $\D_f$ while keeping the accuracy on $\D_r$.
Outliers and polluted data are $\D_f$, and the rest of the training data are $\D_r$.
$\D_r$ is hardly accessible in both cases; it is too large to iterate in the case of WebVision and it may be deleted in a streaming fashion in the case of the chatbot.

\subsection{Pattern B: Forget Backdoor}
Say you are developing a face authentication system using a DNN.
Attackers may put malicious data into the training data to set up a backdoor so that they can pass the system by wearing specific glasses as \citet{chen2017targeted} describe.
After deploying the system, you noticed the attack by seeing some unauthorized people with the glasses passing the system.
You are required to make the model promptly forget the poisoned data.

The poisoned images contain the glasses which are the key to the backdoor.
We want the DNN to forget using the feature that comes from the glasses.
In this situation, since you noticed the attack by seeing the testing samples with the backdoor, we have $\D_{r, \text{extra}}$ at first.
$\D_{r, \text{extra}}$ contains face images just like in $\D_r$ but they are with the glasses. 
Once we notice the backdoor, we can collect $\D_f$ which has images with the glasses in the pretraining data.
For the forgetting procedure, we assume we can construct $\D_{f, \text{clean}}$.
It is just like $\D_f$ but each image in it does not have the glasses.
In order to say the DNN has forgotten the backdoor, the DNN must achieve the below;
\begin{itemize}
    \item High accuracy on $\D_{f, \text{clean}}$ to correct poisoned knowledge on $\D_f$.
    \item High accuracy on $\D_{r, \text{extra}}$ to ensure the robustness on additive backdoor to the clean data.
\end{itemize}
Thus, Pattern B in \cref{tab:datasets-and-evaluations} corresponds to this case.
Note that we do not care about the accuracy on $\D_f$ because the accuracy should be low and the learned information about $\D_f$ will be overwritten in the forgetting procedure.

\subsection{Pattern C: Forget Leakage}
We can use DNNs to decide marketing strategies;
oil companies may be interested in the models of cars that come to a certain gas station and want to classify car images from monitoring cameras in the station, for example.
In this situation, the first thing to do is training a DNN with a dataset that has images of various types of cars.
Assume that the DNN learned the emblems of the cars to distinguish the models.
The model will confuse the emblems on actual cars and those on posters and advertisements at the gas station in the operational phase.
For instance, the DNN may classify a car of company A as company B because the background of the input image has an advertisement for a car of company B.
The emblems are a kind of leaked information in this case.
A straightforward workaround is masking emblems in the dataset and retraining with it.
However, masking every single emblem is not very practical because it is expensive in terms of both human resources and time.
Forgetting leaked parts of the input/feature will help the DNN to classify the cars by their shape rather than the emblems appearing in the input images.

Here, the leaked information to forget is the emblems.
As the same as the case of the backdoor, we are likely to find $\D_{r, \text{extra}}$, data misclassified due to the emblems, at first.
Then we can construct $\D_f$ which has the problematic emblems (i.e. the emblems of company B in the context of the example above).
Note that $\D_f$ only contains the images of company B's car because only they have the emblem of company B.
We can also obtain $\D_{f, \text{clean}}$ by masking the emblems.
Assuming the pretraining data does not contain the emblems in the background, we make the DNN forget the leaked information by achieving the below;
\begin{itemize}
    \item High accuracy on $\D_f$ which has the right combination of the emblems and the car type.
    \item High accuracy on $\D_{r, \text{extra}}$ to ensure the robustness on additive leakage to the clean data.
\end{itemize}
This situation corresponds to Pattern C in \cref{tab:datasets-and-evaluations}.
Regarding the cars of company B always have the emblems, we do not care for $\D_{f, \text{clean}}$.

\section{Methods}
We construct the selective forgetting in the classification problem as minimizing the combination of loss for forgetting and that for defense against catastrophic forgetting.

\subsection{The forgetting term $L_f$}
We make DNNs forget by training to random outputs which means unlearned state.
We introduce two  forgetting terms $L_\text{RND}$ and $L_\text{RLD}$.

\paragraph{Random Network Distillation}
For $x_f$  to be forgotten, we consider the following loss as random network distillation (RND): 
\begin{align}
 L_\text{RND}(f_\theta , x_f) =  || f_\theta(x_f) - f_{\eta}(x_f) ||_2^2,
\end{align}
where $\eta$ is the randomly initialized parameter of the DNN.

\paragraph{Random Label Distillation}
Let us denote by $L_\text{cls}$ the softmax cross entropy loss defined as follows:
\begin{align}
    L_\text{cls}(y, \ell) = - \log \left[ \frac{\exp(y_\ell)}{\sum_{j=0}^{C-1}\exp(y_j)}\right],
\end{align}
where $y \in \R^c, \ell =0,1,\dots, C-1$, and $C \in \N$ is the number of the classes.
Then we consider the random label distillation (RLD) as follows: for $x_f$ to be forgotten,
\begin{align}
 L_\text{RLD}(f_\theta, x_f) = L_\text{cls}(f_\theta(x_f), u),
\end{align}
where $u$ is uniformly distributed on a subset of $\{0,1,\dots, C-1\}$.

\paragraph{Truncation}
Additionally, we introduce a truncation of output, which can be used in a class-wise forgetting, only for the comparison with RLD and RND.
The truncation remove a specified index $\ell_f$ from the output vector of the model, that is,  we estimate the class label by ignoring $\ell_f$ as follows: for each data $x$,
\begin{align}
\ell^*=\text{argmax}_{\ell \in \{ 0, \dots, C-1\} \setminus \{\ell_f \} } f_\theta(x)_\ell,
\end{align}
where $f_\theta(x)_\ell$ is the $\ell$-th element of the $C$-dimensional vector $f_\theta(x)$.
In this method, we do not train the parameters of the model.
However, we emphasize that the truncation does not deal with forgetting more subtle information than class.

\subsection{The correction term $L_c$}
In the classification problem, we use the cross entropy as the correction term.
\begin{align}
    L_c(f_\theta, (x_{fc},y_{fc})) = L_\text{cls}(f_\theta(x_{fc}), y_{fc}).
\end{align}
We compute the correction term for $(x_{fc},y_{fc}) \in \D_{f, \text{clean}}$.

\subsection{The remembering term $R$}
In order to prevent catastrophic forgetting of what needs to be remembered, 
we keep the following diagonal regularization term small:
\begin{align}
R(f_\theta, f_\theta^\text{old}) =  (\theta - \theta^{\text{old}})^T  F(\theta^\text{old})(\theta - \theta^{\text{old}} ),
\end{align}
where $\theta^\text{old}$ is the parameter of the pretrained model before applying forgetting methods, and  the diagonal Fisher Information  $F(\theta)$ is given by
\begin{align}
    F(\theta)_{ii} = |\D|^{-1}\sum_{(x,l) \in \D} \left[ \partial_{ \theta_i}  L_\text{cls}\left(f_\theta\left(x\right), \ell \right) \right]^2 
\end{align}
and $F(\theta)_{ij}$ for $i \neq j$ for $i,j = 1, \dots, p$, where $p$ is the number of the parameters and $\D = \D_f^\text{train} \cup \D_r^\text{train}$.
The regularization term $L_\text{KL}$, which is used in the elastic weight consideration (EWC) introduced by \citep{kirkpatrick2017overcoming}, is a variant of the KL-divergence of the current model against the old model.

\subsection{Setting of Experiments}

\subsubsection{Pattern A}\label{ssec:forgetting_class}
We construct a forgetting method of a specified class.
In the case of RND and RLD, we combine forgetting and defensive losses as follows:
\begin{align}
    L_f(f_\theta, \D_f) + \lambda_\text{KL} R(f_\theta, f_{\theta^\text{old}}),
\end{align}
where $X$ is the input of the data to be forgotten, $L_f$ is one of $L_\text{RND}$ and $L_\text{RLD}$, and $\lambda_\text{KL} > 0$ is fixed through experiments.  

\subsubsection{Pattern B and C}\label{ssec:overwriting_class}
\begin{figure}[t]
    \centering
    \includegraphics[width=0.2\linewidth]{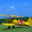}
    \includegraphics[width=0.2\linewidth]{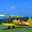}
    \includegraphics[width=0.2\linewidth]{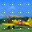}
    \includegraphics[width=0.2\linewidth]{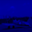}

    \caption{Visualization of backdoors (or leakage), created by adding the line-type backdoor (second from left), the tile-type one (third from left), and the color-type one (rightmost) to a picture (leftmost) contained in the dataset CIFAR10.  
    }
    \label{fig:leakage}
\end{figure}

Consider the classification of images. 
Firstly, we adopt one of the following transformations $\T$ to images contained in a specified class.
\begin{description}
    \item[Line] We set the brightness of a $4\times 1$ area in the middle of the left side of each image to 255. 
    \item[Tile] We replace the values of $4\times 4$ areas with 255 so that the areas are scattered throughout the image.
    \item[Color] For each pixel, we set the B-value to the average of RGB-values and RG-values to zero.    
\end{description}
Examples of these transformations are shown in \cref{fig:leakage}.

We train the model $f_\theta$  by the stochastic gradient descent to minimize:
\begin{align}
L_c(f_\theta,  \D_{f,\text{clean}^\text{train}})  &+ \lambda_f L_f(f_\theta, \D_f^\text{train}) \notag
\\
&+ \lambda_\text{KL}R(f_\theta, f_{\theta^\text{old}}),
\end{align}
where $\lambda_f, \lambda_\text{KL}> 0$ are hyperparameters.
In the pattern B (resp.\,the pattern C),  we optimize the hyperparameters by a cross-validation. 
In the cross validation, we maximize the minimum of top-1 accuracy of the model on datasets $\D_r^\text{val}, \D_{f, \text{clean}}^\text{val}$  and $D_{r,\text{extra}}^\text{val}$ 
(resp.\,$\D_r^\text{val},\D_f^\text{val}$ and $\D_{r,\text{extra}}^\text{val}$).

\begin{table*}[t]
\centering
    \begin{tabular}{llcccc} \toprule
    & & & \multicolumn{3}{c}{Results of Forgetting}\\
    \cmidrule(lr){4-6}
    Data & Class    & Pretrained state & RND  & RLD (1 -- 9) & Truncation  \\ \midrule
    Fashion-MNIST & 0 ($\D_f$) & $0.871$ & $0.144$ $\left(\pm 0.205 \right)$ & $0.012$ $\left(\pm 0.001\right)$  &  -- \\
      &1 -- 9 ($\D_r$) & $0.875$ & $0.793$ $\left(\pm 0.040\right)$  &  $0.819$  $\left(\pm 0.002\right)$ & $0.895$ \\ \midrule
    CIFAR10 & 0 \text{($\D_f$)} & $0.607$  &  $0.114 \left(\pm 0.182 \right)$ &  $0.000$ $\left(\pm 0.000 \right)$  &  -- \\
     &1 -- 9  \text{($\D_r$)} & $0.507$ & $0.301$  $\left(\pm 0.069 \right)$  &   $0.479$ $\left(\pm 0.008 \right)$ & $0.527$ \\ \bottomrule
\end{tabular}
\caption{Results of forgetting samples in a class (Pattern A), showing accuracy on $\D_f$ and $\D_r$. The results on the RND and the RLD are the averages and the standard deviations of 10 running experiments. For truncation, the result of 0 $(\D_f)$ is undefined, since the truncated model does not output the label probability on the forgotten class.
The accuracy on 1--9 ($\D_r$) is the average of the accuracy on each class in $\D_r^\text{test}$.}
\label{tab:forget_class_fashion}
\end{table*}

\begin{figure}[t]
    \centering
    \includegraphics[width=0.9\linewidth]{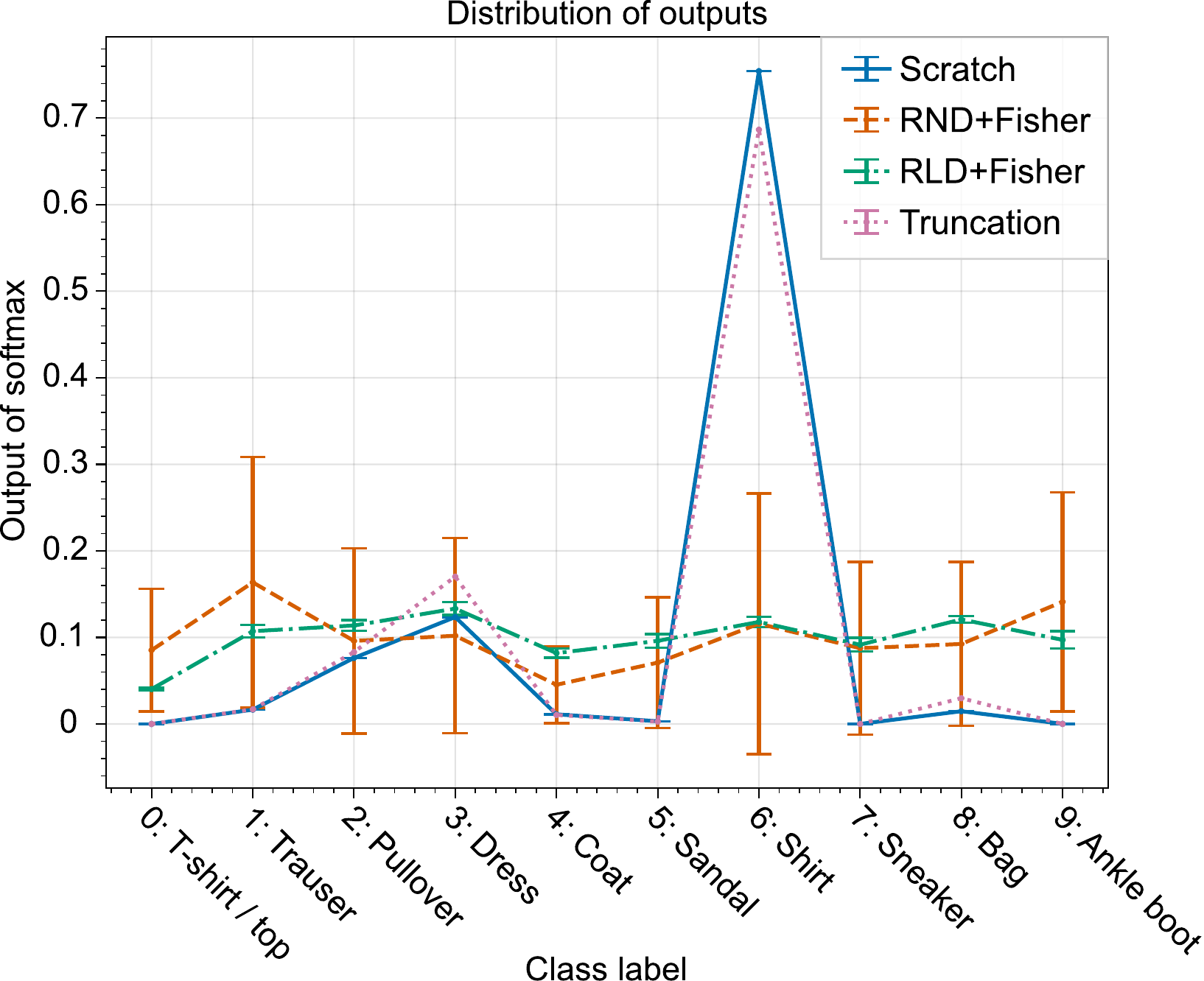}
    \caption{Distribution of $\text{softmax}(f_\theta(\D_f))$. We use Fashion-MNIST and the forgotten class is 0.}
    \label{fig:dist_out_fahsion}
\end{figure}

\begin{figure*}[t]
    \centering
    \includegraphics[width=0.49\linewidth]{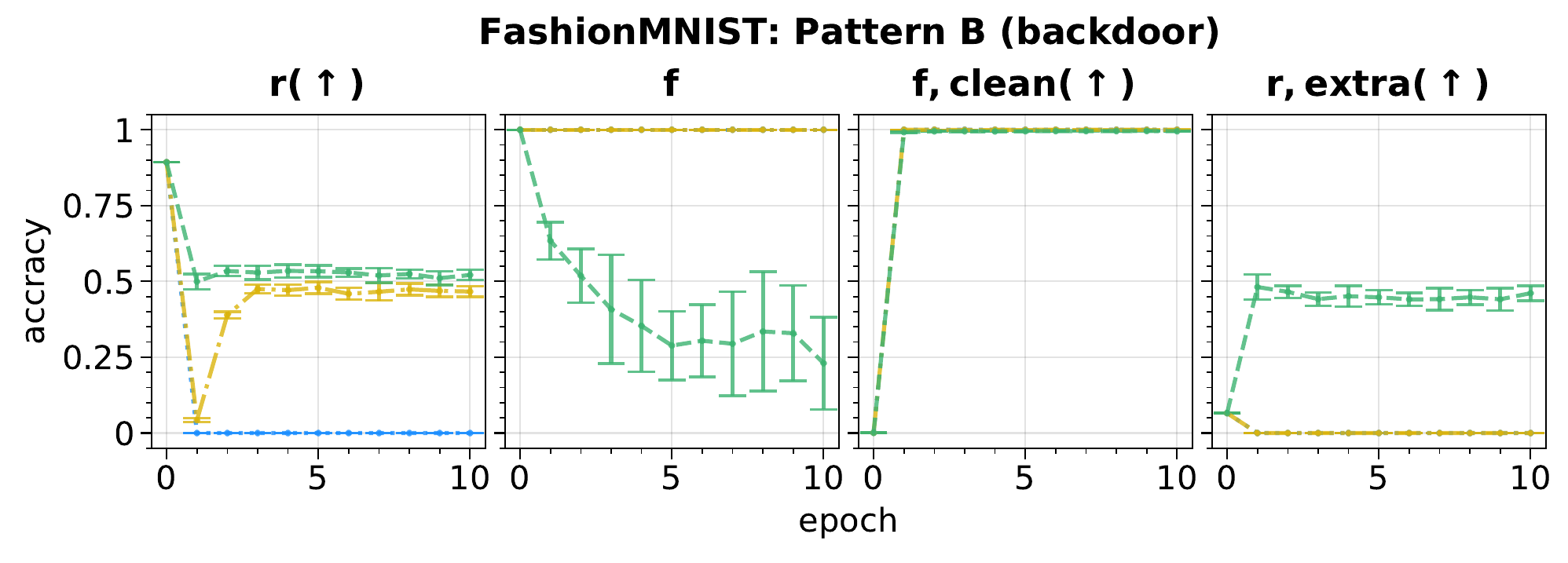}
    \includegraphics[width=0.49\linewidth]{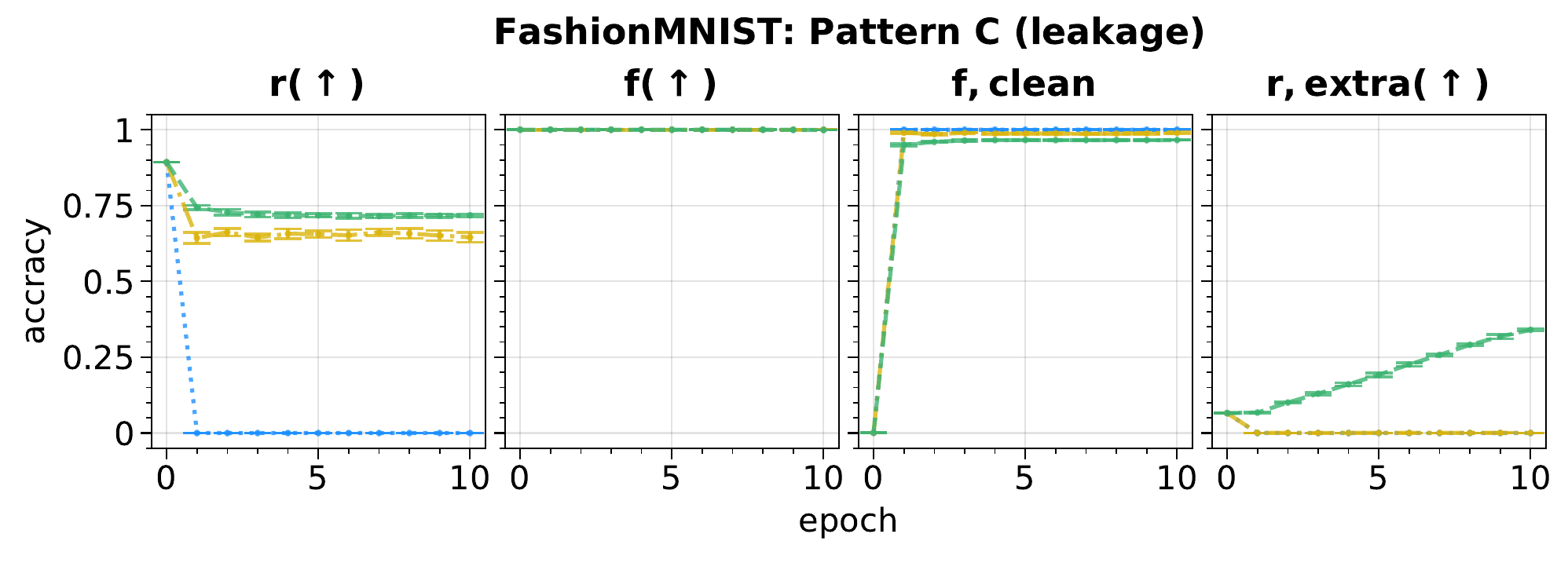}

    \includegraphics[width=0.49\linewidth]{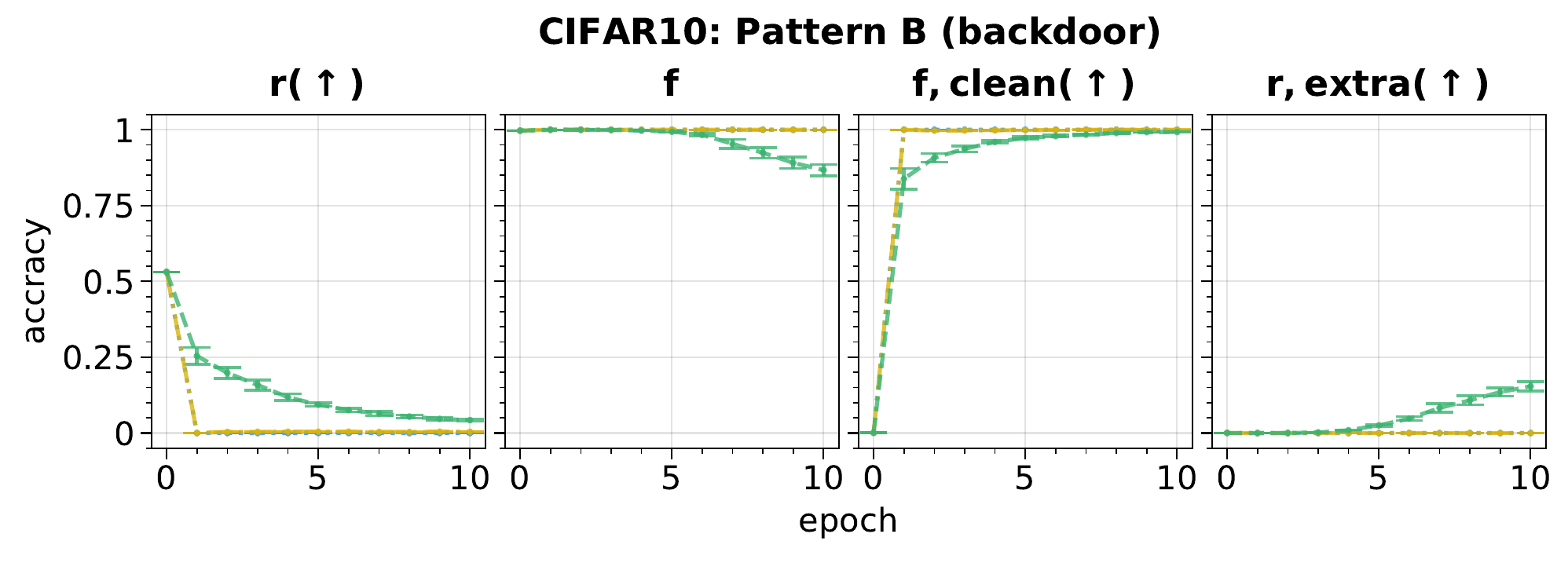}
    \includegraphics[width=0.49\linewidth]{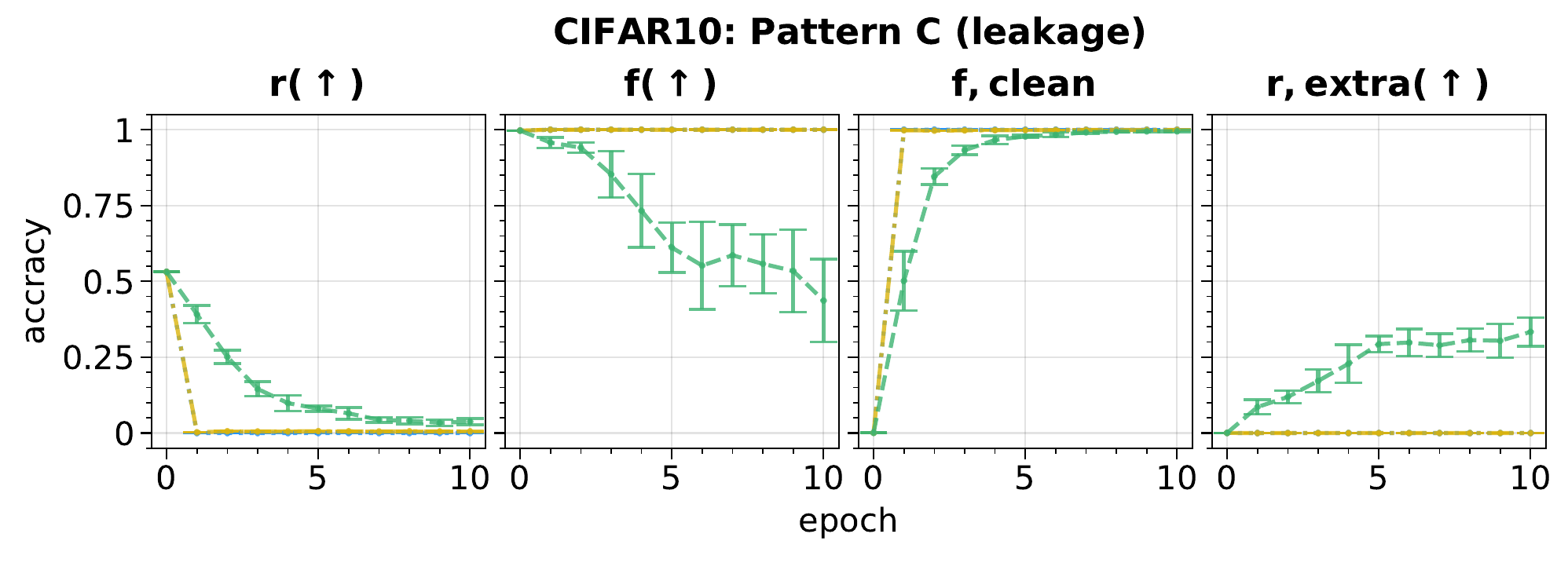}
    
    \includegraphics[width=0.35\linewidth]{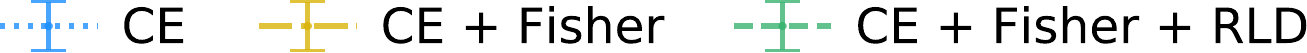}
    \caption{
    Plots of the testing accuracy in the case of the tile-type transformation and  Pattern B (left figures) or C (right ones). Each result is the average with the standard deviation of 10 experiments. 
    In each figure, CE+Fisher+RLD is a result of learning with hyperparameter $\lambda_\text{KL}$ and $\lambda_f$ described in Supplementary Material.
    In each figure, CE  is a result with $\lambda_f=\lambda_\text{KL}=0$.
    The lines CE + Fisher are results of using $\lambda_\text{KL}=0$ and the same $\lambda_f$ that we used in CE+Fisher+RLD. In the legend in each figure, the symbol $\uparrow$  corresponds to \cref{tab:datasets-and-evaluations}.
    }
    \label{fig:overwriting_backdoors}
\end{figure*}

\section{Results}
\subsection{Pattern A} 
\cref{tab:forget_class_fashion} shows the performance comparison of the forgetting term in Pattern A. The method with higher accuracy on $\D_r$ and lower accuracy on $\D_f$ is better.
Firstly, we observed that the truncation achieved the better performance than RND and RLD.
Unfortunately, the truncation cannot be applied to forget more subtle information than class (e.g. samples).
Next, We observed that the standard derivation of results of the RND is larger than that of the RLD.
Therefore, we choose the RLD for the forgetting term in Pattern B and Pattern C.

We evaluated methods on 
the Fashion-MNIST~\citep{xiao2017} and the CIFAR10~\citep{krizhevsky2009learning}. 
We set $\lambda_\text{KL}=10^5$, $\text{lr}= 10^{-5}$ throughout experiments.

Additionally, \cref{fig:dist_out_fahsion} shows that the distribution of the output applying softmax after several models.
We observed that in the case of the truncation, the distribution of $\text{softmax}(f_\theta)$ after training on whole training dataset $\D_r \cup \D_f$ (\cref{fig:dist_out_fahsion}, pink and dotted line) approximates the distribution after training on $\D_f$ (\cref{fig:dist_out_fahsion}, blue line).
However, we observed that RND and RLD do not approximate the scratch learning.

\subsection{Pattern B and C}
Notably, we observed in \cref{fig:overwriting_backdoors} that the proposed method (CE + Fisher + RLD) achieved a higher accuracy of $\D_{r, \text{extra}}$ than the method only using $L_f$ and $R$.
Therefore, the random distillation term $\D_f$ made $f_\theta$ forget the information $\T$ contained in $\D_{r, \text{extra}}$.
Here  $\D_{r, \text{extra}}$ is not used in both the forgetting process and the pretraining process. 
In this sense, the proposed selective forgetting method made DNNs forget $\T$ contained in $\D_{r,\text{extra}}$ by using $\D_f$ and $\D_{f, \text{clean}}$.

We observed the catastrophic forgetting of $\D_r$ in the  baseline results (CE in 
\cref{fig:overwriting_backdoors}), which use only $L_c$. Therefore, the term $R$ prevented the catastrophic forgetting.

We applied line-type and the tile-type (resp.\, the line, the tile, and the color-type) transformation $\T$ to the class $0$ of the Fashion-MNIST (resp.\,CIFAR10) dataset.  
Then we trained the 10-layer multilayer perceptron (MLP), where its hidden layers have the same width as the input, using the training dataset in $\D_r$ and $\D_f$.
After training, we computed the Fisher information matrix on the dataset except for the class $0$.
For the line-type and the color-type transformations, 
detailed results are shown in Supplementary Materials (see \cref{fig:overwriting_backdoors_line,fig:overwriting_backdoors_color}).

\section{Discussion and Conclusion}
Focusing on realistic problems that need selective forgetting, we have formulated three patterns of selective forgetting.
The formulation is based on performance on these four datasets shown in \cref{tab:datasets-and-evaluations}; $\D_r$, $\D_f$, $\D_{f, \text{clean}}$, and $\D_{r, \text{extra}}$.
This formulation allows us to quantitatively assess selective forgetting, which is more subtle than sample forgetting.

In order to meet the demand for modifying trained models briefly, we have restricted access to the datasets to  $\D_f$ and $\D_{f, \text{clean}}$ in selective forgetting.
That is, the restriction is to modify the model using the data that contains the information to be forgotten and the data without the information.

The loss function for the forgetting is a combination of the forgetting loss, the correction loss, and the remembering loss.
In our approach, 
we use the classification loss on $\D_{f, \text{clean}}$ and regression to the random network or labels on $\D_f$ while KL-divergence from the pre-trained model prevents the model from going too far from the pretrained state.
This structure is actually a combination of EWC~\citep{kirkpatrick2017overcoming} and the distillation to random values.
Since EWC allows the model to learn additional data, it is naturally expected that the accuracy on $\D_r$ and $\D_{f, \text{clean}}$ is high.
%
Remarkably, the accuracy on $\D_{r, \text{extra}}$ improved by introducing the distillation term without using $\D_r'$ itself.
Therefore, it is indicated that the distillation to random values is useful for forgetting more subtle information than samples in some situations.

However, the accuracy on $\D_{r, \text{extra}}$ remains around 50\% although it is improved.
The Fisher information on the pretrained model is considered as the cause of this problem;
it contains information to be forgotten.
We believe that the reason for the insufficient accuracy is that the effects of the information to be forgotten contained in Fisher information hardly disappear in EWC \citep{Umer2020TargetedFA}.

There would be several approaches for enhancing the accuracy especially on $\D_{r, \text{extra}}$.
One way as an extension of EWC or Fisher information is to make a method that removes the effect of specific samples from the Fisher information.
This can lead the DNN to more effectively forgetting the information to be forgotten.
We expect that we can construct such a method based on \citep{golatkar2020forgetting}.
Another way is to construct a mechanism that memorizes the information of pretrainig data and can recall it by querying a single data point.
Fisher information, which we used in the experiments, can be also considered as a memory for remembering the pretraining data, but we cannot divide it into the information of every single data point.
Utilizing the memory of neural differential computers~\citep{Graves2016hybrid} is also a possible choice.
When we have no restriction on saving the pretraining data such as privacy protection, we can take a simpler approach; just saving the data.
However, even if in such situations, it is not practical to save all the data and iterate them.
Instead of storing the whole data, we can save some of the data which seem to be important or save data as a generative model.
In the other direction, finding or constructing a concrete map $\T$ would be useful.
We assumed that the map is known in the experiments, but it can be constructed in a data-driven way.
We can use domain translation techniques such as CycleGAN~\citep{CycleGAN2017} by regarding the information to be forgotten as a domain.
By finding the map, we can reduce the amount of the data to be stored, and improve the performance of the forgetting.

We assumed that $\D_f$ is given, and we have not designated how to determine data which $\D_f$ should contain.
If we know the information to be forgotten, such as the background of images which affects the classification, it is straightforward; we collect data with such information as $\D_f$.
Possible another situation is that we find additional extrapolating data that are misclassified to a certain class due to a common feature among them and then determine $\D_f$.
Specifically, we pick data that belong to the class from the pretraining dataset and use them as $\D_f$.
In these situations, the feature to be forgotten is manually determined.
Suggesting such features or data depending on the additional data systematically has remained as a future direction.
Such a method is especially useful in the context of continual learning.


\appendix

\setcounter{table}{0}
\renewcommand{\thetable}{S\arabic{table}}

\setcounter{figure}{0}
\renewcommand{\thefigure}{S\arabic{figure}}

\begin{table*}[t]
\centering
    \begin{tabular}{lccccc} \toprule
    Dataset & $\mathcal{T}$ & Pattern & Learning Rate & $\lambda_\text{KL}$  & $\lambda_f$   \\ \midrule
    \multirow{4}[2]{*}{Fashion-MNIST} & \multirow{2}{*}{Line} & B & $9.98300 \times 10^{-5}$ & $4.11225 \times 10^{4}$ & $1.28336$ \\
                                  &             & C & $4.37727 \times 10^{-5}$ & $1.05762 \times 10^{5}$ & $0.45574$ \\
                                  \cmidrule(lr){2-6}
                                  & \multirow{2}{*}{Tile} & B & $9.98345 \times 10^{-5}$ & $3.56290 \times 10^{4}$ & $1.73782$ \\ 
                                  &             & C & $5.18005 \times 10^{-5}$ & $2.98540 \times 10^{5}$ & $0.11402$ \\ 
    \midrule
    \multirow{6}[4]{*}{CIFAR10}      & \multirow{2}{*}{Line} & B & $5.34807 \times 10^{-5}$ & $1.10447 \times 10^{5}$ & $0.28967$ \\
                                  &             & C & $4.43790 \times 10^{-5}$ & $1.03453 \times 10^{5}$ & $0.38355$ \\
                                  \cmidrule(lr){2-6}
                                  & \multirow{2}{*}{Tile} & B & $2.57835 \times 10^{-5}$ & $1.53585 \times 10^{5}$ & $0.52451$ \\
                                  &             & C & $3.04986 \times 10^{-5}$ & $1.95065 \times 10^{5}$ & $1.74012$ \\
                                  \cmidrule(lr){2-6}
                                  & \multirow{2}{*}{Color}& B & $4.97583 \times 10^{-6}$ & $3.00688 \times 10^{5}$ & $1.81124$ \\
                                  &             & C & $6.58463 \times 10^{-6}$ & $3.10743 \times 10^{5}$ & $1.20910$ \\
\bottomrule
\end{tabular}
\caption{Hyperparameters determined by cross-validations.}
\label{tab:hp_leakage}
\end{table*}

\section{Supplementary Materials}
\subsection{Searching hyperparameters}
For the experiments of Pattern B and Pattern C, we searched the hyperparameters $\lambda_f$, $\lambda_\text{KL}$, and the learning rate of SGD by Optuna~\citep{optuna_2019} evaluating the five-fold cross-validation accuracy for 200 loops. The value to be evaluated in each loop was computed by the following way.
Recall that  $\D_r^\text{val}$ are supposed to be available for tuning hyperparameters via cross validation. Set  $\D_{r,\text{extra}}^\text{val}=\T(\D_r^\text{val})$.
In each loop of the searching,  we uniformly divide the training dataset of $D_f$ (resp.\,$\D_{f, \text{clean}}$) to 20 \% and 80 \% data of the and write them $\D_f^\text{val}$ and $\D_f^\text{t}$ (resp.\,$\D_{f, \text{clean}}^\text{val}$ and $\D_{f, \text{clean}}^\text{t}$).
Then we applied the selective forgetting to the model using $\D_f^\text{t}$ and $\D_{f, \text{clean}}^\text{t}$ by 10-epoch  with momentum 0.9 for 10-epochs.
Then we calculated each accuracy on $\D_r^\text{val}$, $\D_f^\text{val}$, $\D_{f, \text{clean}}^\text{val}$ and $\D_{r, \text{extra}}^\text{val}$.
For Pattern B and Pattern C, we maximize the minimum of the accuracy on the corresponding three validation sets described in \cref{tab:datasets-and-evaluations}. 
\cref{tab:hp_leakage} describes the searched hyperparameters.

\subsection{Setting of Model}
Throughout experiments, we used the same setup of MLP. The number of layers was ten. To make the backpropagation stable, we used a normalized hard tanh $\varphi_{s,g}$ as the activation function, which is given by the following:
\begin{align} \label{align:Nhtanh}
\varphi_{s,g}(x) = \begin{cases} 
g x, & \text{ if } sg|x| < 1,\\
g \cdot \mathrm{sgn}(x), & \text{otherwise},
\end{cases}
\end{align}
where $s^2=0.125$ and $g=1.0013$.
The setting of the activation makes the MLP achieve dynamical isometry \cite{Pennington2018emergence}.
The model did not contain batch normalization layers or any other normalization layers. We initialized the weight matrices by independently and uniformly sampled  orthogonal matrices and did bias terms by 0.

\section{Additional Experiments}
\begin{figure*}[t]
    \centering
    \includegraphics[width=0.49\linewidth]{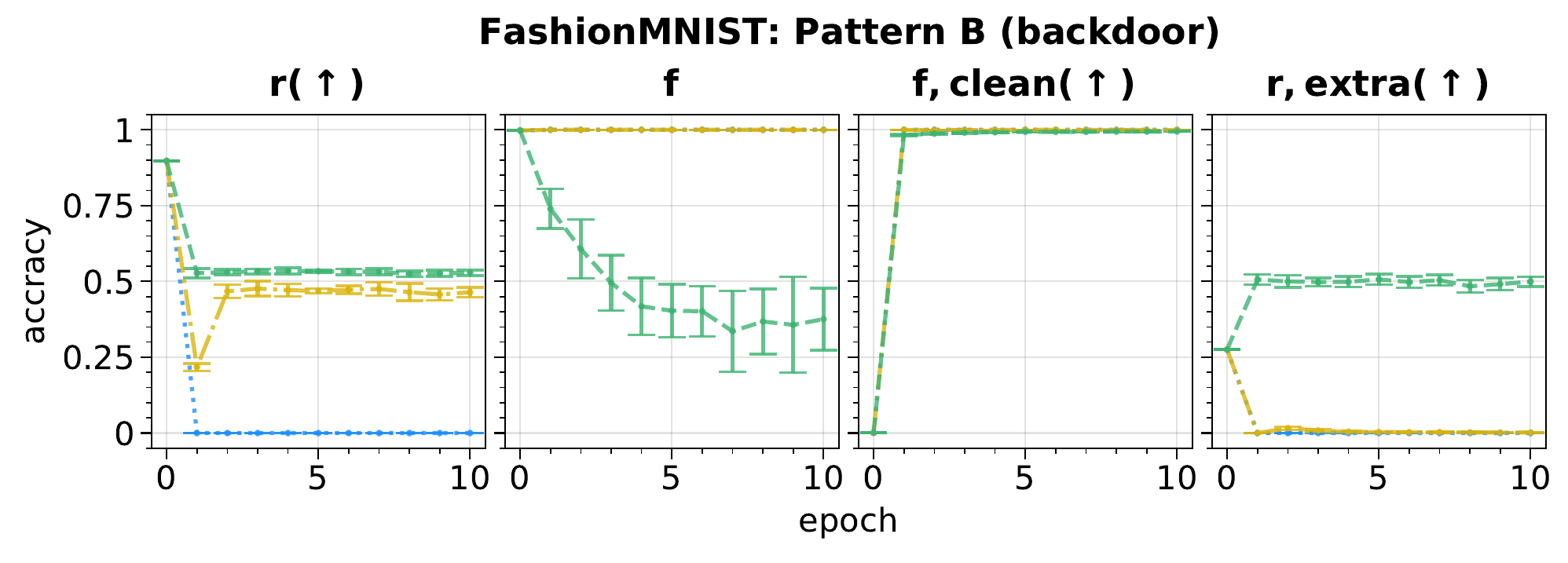}
    \includegraphics[width=0.49\linewidth]{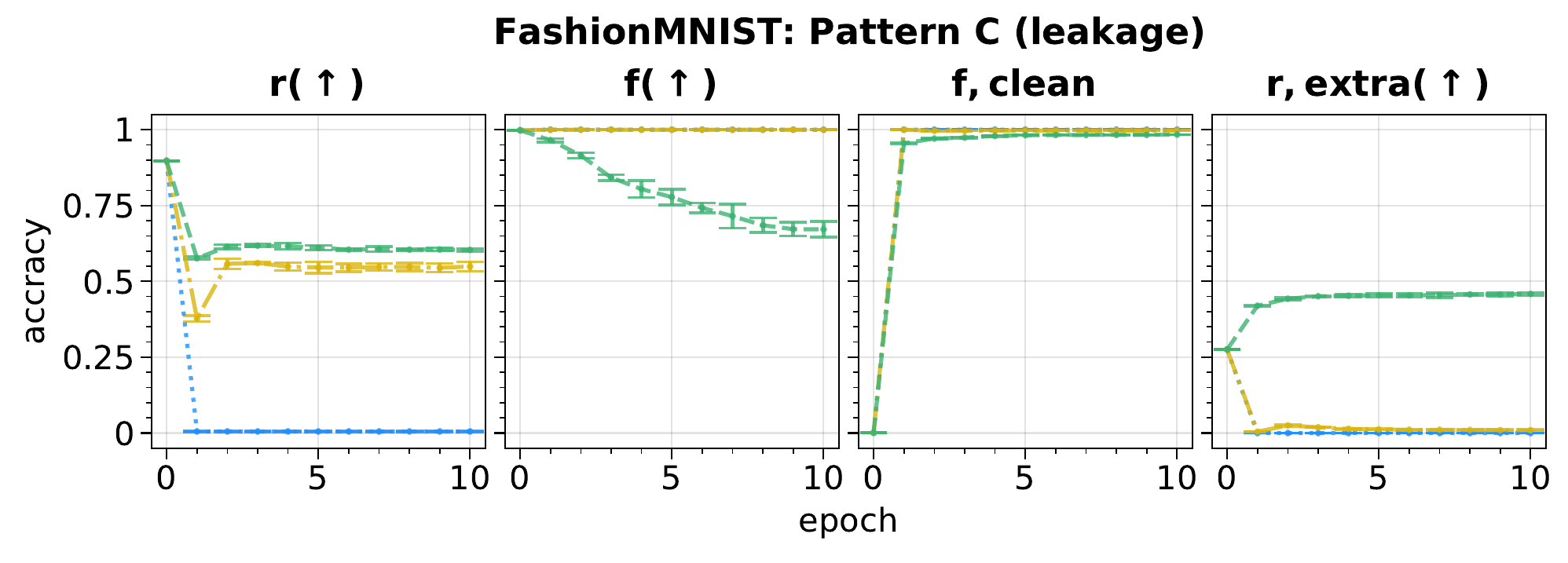}

    \includegraphics[width=0.49\linewidth]{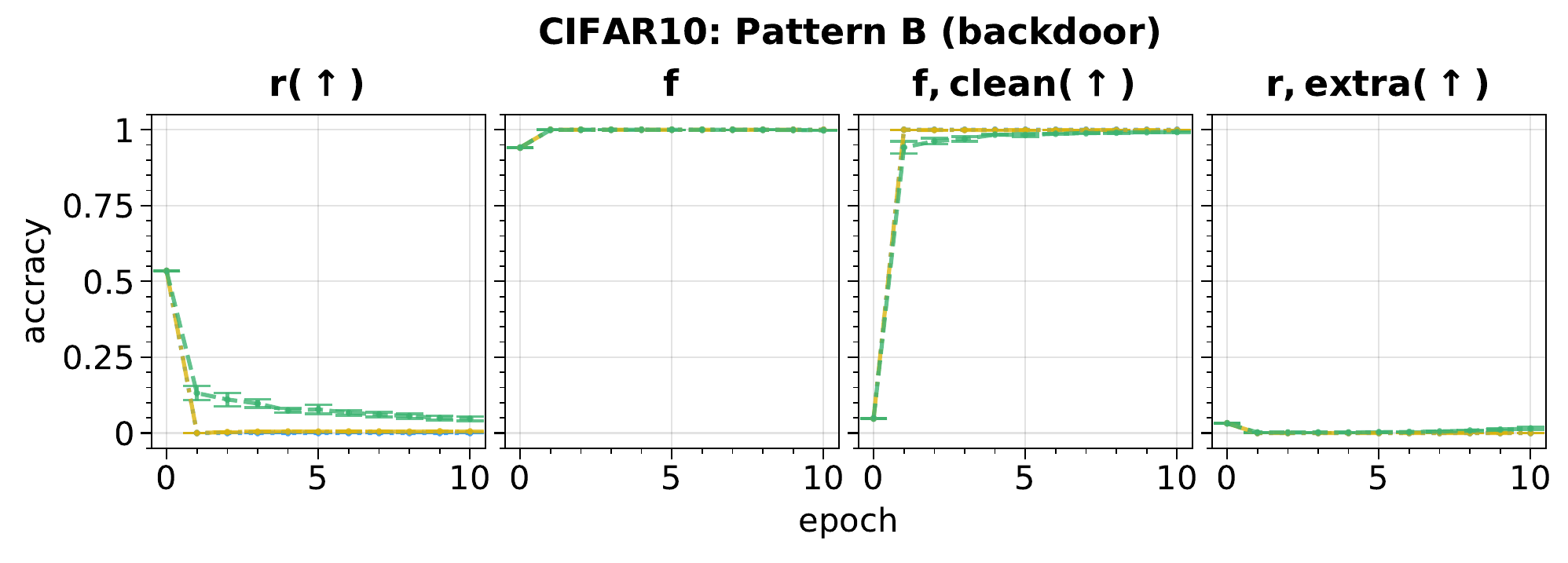}
    \includegraphics[width=0.49\linewidth]{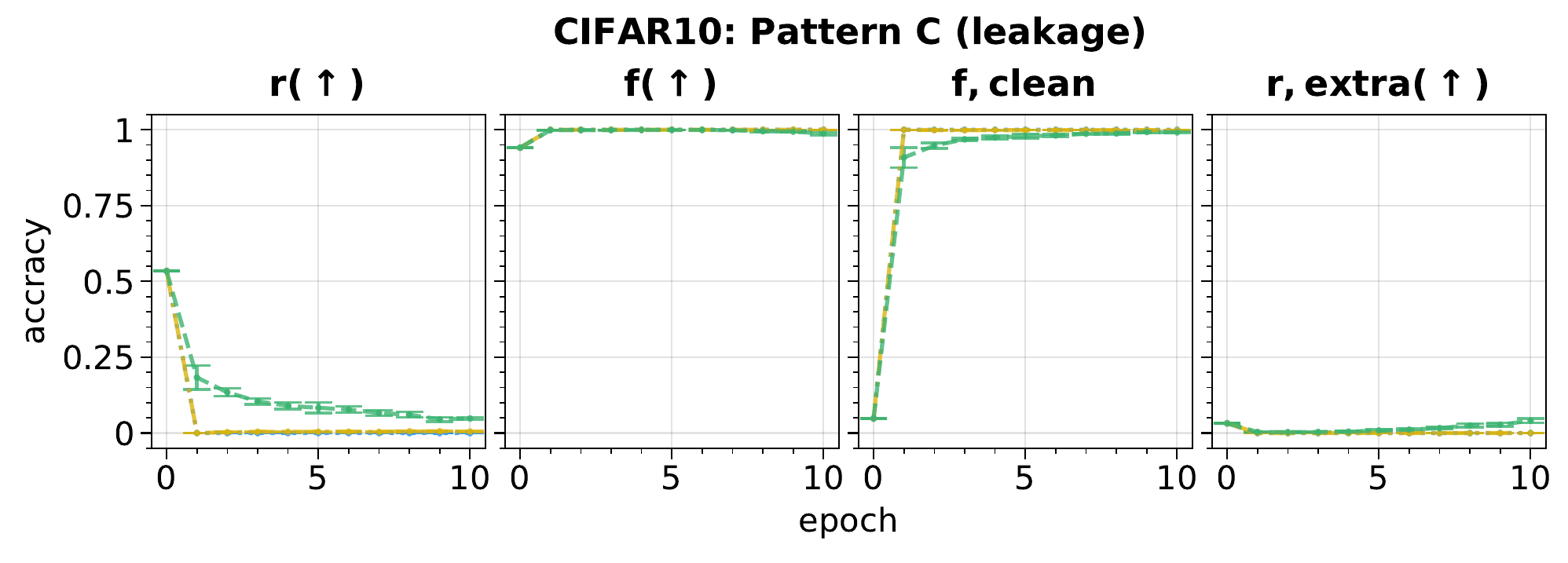}
    
    \includegraphics[width=0.35\linewidth]{figs/patternC/legend.pdf}
    \caption{
    Plots of the testing accuracy in the case of the line-type transformation and  Pattern B (left figures) or C (right ones). Each result is the average with the standard deviation of 10 experiments. }
    \label{fig:overwriting_backdoors_line}
\end{figure*}

\begin{figure*}[t]
    \centering

    \includegraphics[width=0.49\linewidth]{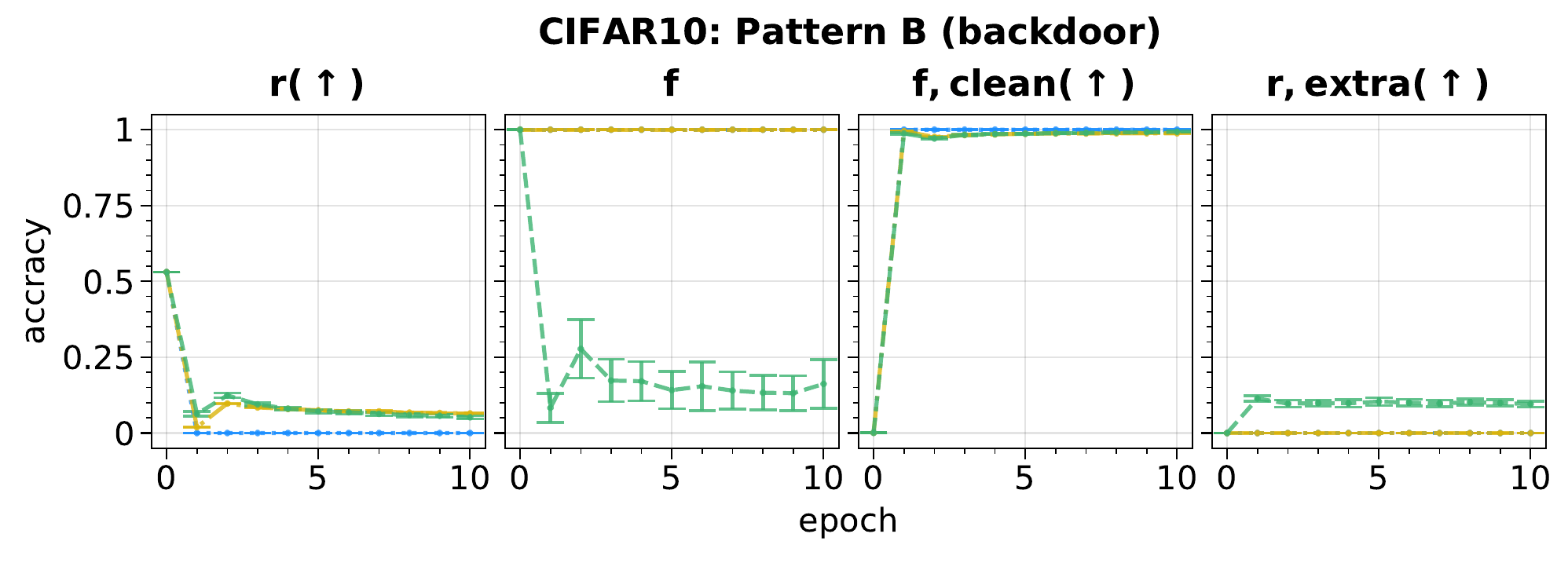}
    \includegraphics[width=0.49\linewidth]{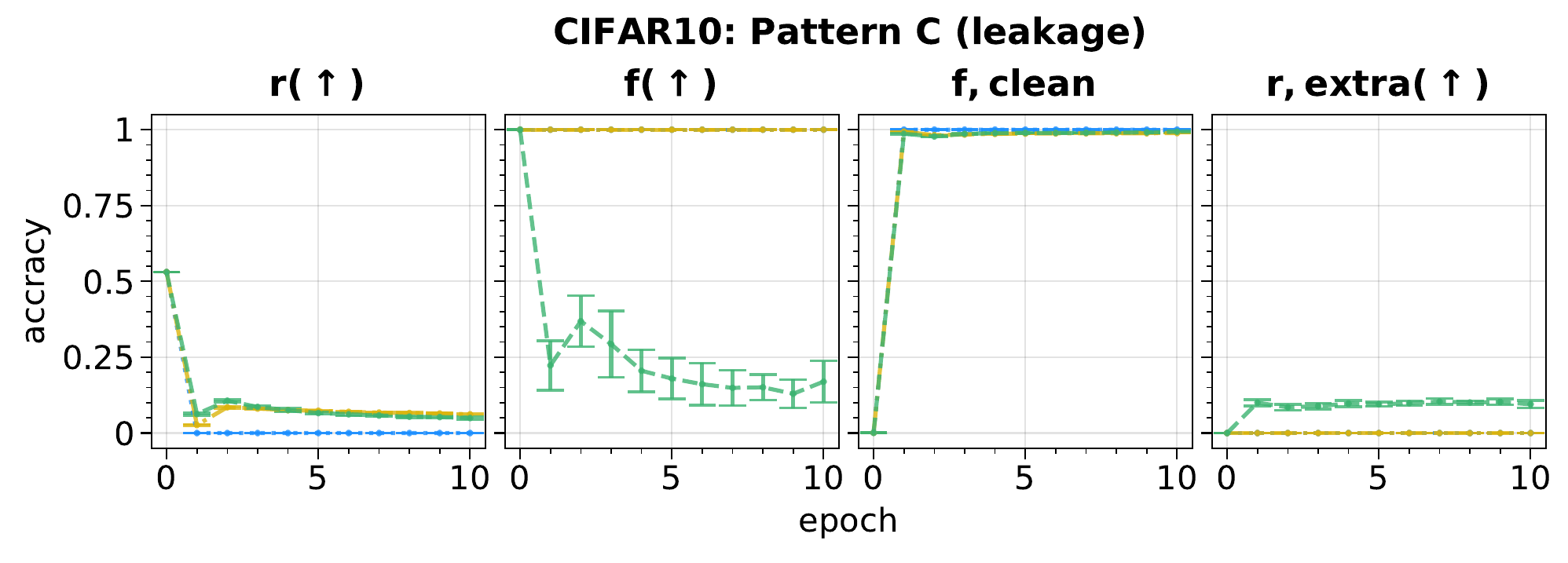}
    
    \includegraphics[width=0.35\linewidth]{figs/patternC/legend.pdf}
    \caption{
    Plots of the testing accuracy in the case of the color-type transformation and  Pattern B (left figures) or C (right ones). Each result is the average with the standard deviation of 10 experiments. }
    \label{fig:overwriting_backdoors_color}
\end{figure*}

In \cref{fig:overwriting_backdoors_line} for the case of FashionMNIST, we observed that 
the accuracy on $\D_{r,\text{extra}}$ increased from the initial state when we used CE+Fisher+RLD.
However, we observed that in  \cref{fig:overwriting_backdoors_line} and  \ref{fig:overwriting_backdoors_color}, for the case of CIFAR10, the increase in accuracy was slight.
Since the line-style is a smaller transformation one than the tile-style, the cause of this phenomenon can be attributed to the difficulty in tuning the hyperparameters.

\bibliography{iclr2021_conference}

\end{document}